\documentclass{article}
\usepackage{spconf,amsmath,graphicx,epsfig}
\usepackage{amssymb}
\usepackage{booktabs}
\usepackage{multirow}
\usepackage{algorithm}
\usepackage{algorithmic}
\usepackage{bm}
\usepackage{hyperref}
\usepackage{colortbl}
\usepackage{setspace}
\usepackage{subcaption}
\usepackage{cleveref} 
\usepackage{enumitem}
\setlist[itemize]{leftmargin=*}
\hypersetup{colorlinks=true,
            linkcolor=red,
            citecolor=black}

\def\ul#1{\underline{#1}}

\title{SEMANTIC ENHANCED FEW-SHOT OBJECT DETECTION}

\name{Zheng Wang$^{1}$\thanks{$^{\star}$ denotes corresponding author (qingjie.liu@buaa.edu.cn). This work was supported by the National Natural Science Foundation of China under Grant 62176017. \newline
© 20XX IEEE. Personal use of this material is permitted. Permission from IEEE must be obtained for all other uses, in any current or future media, including reprinting/republishing this material for advertising or promotional purposes, creating new collective works, for resale or redistribution to servers or lists, or reuse of any copyrighted component of this work in other works.},
      Yingjie Gao$^{1}$,
      Qingjie Liu$^{1,2}$$^{^\star}$,
      Yunhong Wang$^{1,2}$}
\address{$^{1}$ State Key Laboratory of Virtual Reality Technology and Systems, Beihang University, Beijing, China\\
$^{2}$ Hangzhou Innovation Institute, Beihang University
}

\begin{document}

%
\maketitle
\begin{abstract}
Few-shot object detection~(FSOD), which aims to detect novel objects with limited annotated instances, has made significant progress in recent years. However, existing methods still suffer from biased representations, especially for novel classes in extremely low-shot scenarios. During fine-tuning, a novel class may exploit knowledge from similar base classes to construct its own feature distribution, leading to classification confusion and performance degradation. 
To address these challenges, we propose a fine-tuning based FSOD framework that utilizes semantic embeddings for better detection. In our proposed method, we align the visual features with class name embeddings and replace the linear classifier with our semantic similarity classifier. Our method trains each region proposal to converge to the corresponding class embedding. Furthermore, we introduce a multimodal feature fusion to augment the vision-language communication, enabling a novel class to draw support explicitly from well-trained similar base classes.
To prevent class confusion, we propose a semantic-aware max-margin loss, which adaptively applies a margin beyond similar classes. As a result, our method allows each novel class to construct a compact feature space without being confused with similar base classes.
Extensive experiments on Pascal VOC and MS COCO demonstrate the superiority of our method. 
\end{abstract}

\begin{figure}[t]
    \centering
    \includegraphics[width=\linewidth]{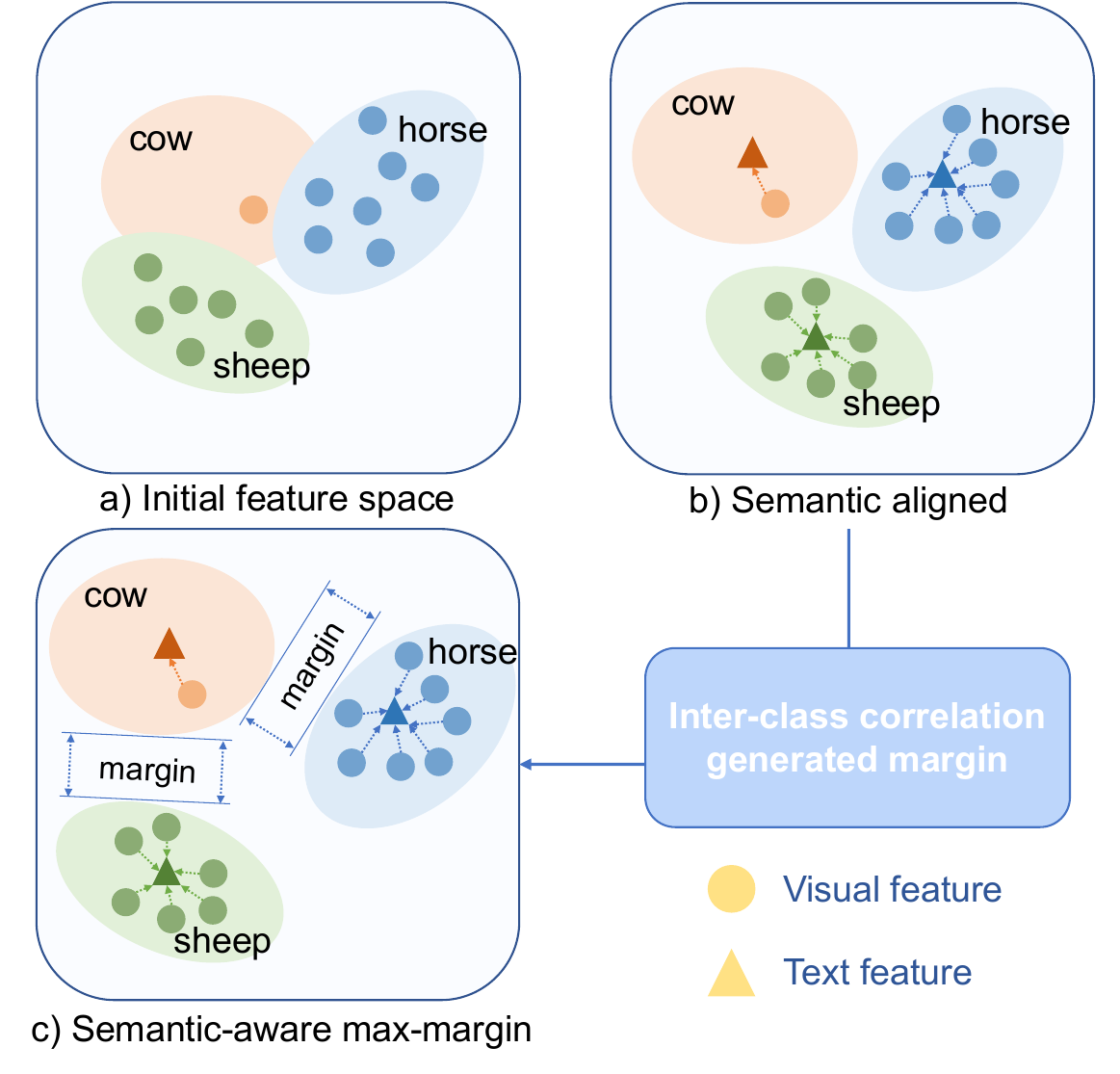}
    \caption{The illustration of our main idea. a)~The original feature space, where 'cow' is a novel class, 'sheep' and 'horse' are confusable base classes. b)~The semantic alignment learning aligns visual space with semantic space by bringing RoI features closer to their class name embeddings. c)~A max-margin loss is further proposed to push confusable classes away from each other.}
    \label{fig:core_idea}
    
\end{figure}

\section{Introduction}
\label{sec:intro}

Deep neural networks have made great strides in object detection recently. However, deep detectors demand a large amount of annotated data to effectively recognize an object. A human, on the other hand, only needs a few samples to identify a class of objects.
General detectors suffer from over-fitting in few-shot scenarios. Closing the performance gap between general detection and detection in few-shot scenarios has become a key area of interest in the computer vision community.

Compared to few-shot classification and general object detection, few-shot object detection~(FSOD) is a much more challenging task. Given base categories with sufficient amounts of data and novel categories with only a few labeled bounding boxes, FSOD focuses on learning basic knowledge on base classes and generalizing well on novel classes. Early FSOD methods prefer to follow the meta-learning paradigm to learn task-agnostic knowledge and quickly adapt to novel tasks~\cite{iccv19_fsrw, iccv19_metarcnn, iccv21_qafewshot, aaai22_metafaster}. However, these methods require a complex training process and often result in unsatisfactory performance in realistic settings. On the other hand, fine-tuning based methods adopt a simple yet effective two-stage training strategy and achieve comparable results~\cite{icml20_tfa, cvpr21_fsce, iccv21_defrcn}. 

In recent years, many studies have focused on fine-tuning based FSOD, which aims to transfer knowledge learned from abundant base data to novel categories. 
TFA~\cite{icml20_tfa} reveals the potential in simply freezing the last layers during fine-tuning, laying the foundation for fine-tuning based approaches. DeFRCN~\cite{iccv21_defrcn} decouples classification and regression by scaling and truncating gradients and achieves superior performance.
Despite their success, there still exists two underlying problems.

First, previous fine-tuning based FSOD methods suffer from performance degradation when training samples are extremely limited, for example, when there is only one annotated box for each category. It is reasonable that only one object cannot well represent a category with diverse appearance. 
The biased representations severely damage the performance of novel classes.
Second, FSOD performance continues to be threatened by confusion between novel and base categories. With only few annotated samples, a novel class can hardly construct a compact feature space. As a result, a novel class may scatter in well-constructed feature spaces of similar base classes, leading to confusion in classification.


In this work, we propose a fine-tuning based framework which utilizes semantic embeddings to improve generalization on novel classes. The framework exploits the semantic information implied in the class names and replaces the linear classifier with a Semantic Similarity Classifier~(SSC) in the novel fine-tuning stage. The SSC produces classification results by calculating cosine similarity between class name embeddings and region features of proposals. Further, a Multimodal Feature Fusion~(MFF) is proposed to perform deep fusion of visual and textual features. We also apply Semantic-Aware Maxmargin~(SAM) loss upon the original cross entropy loss to separate novel categories and base categories that resemble themselves, as illustrated in Fig.~\ref{fig:core_idea} During fine-tuning, the SSC and MFF are optimized in an end-to-end manner by classic Faster R-CNN~\cite{nips15_fasterrcnn} loss and the SAM loss. The contributions of this work can be summarized as follows:

\begin{itemize}
    \setlength{\itemsep}{0pt}
    \setlength{\parsep}{0pt}
    \setlength{\parskip}{0pt}
    \item We propose a framework that utilizes semantic information to address issues related to low-shot performance degradation and class confusion.
    \item In order to tackle these issues, we further design three new modules, i.e. SSC, MFF and SAM loss, which provide unbiased representations and increase inter-class separation.
    \item Extensive experiments on both PASCAL VOC~\cite{dataset_voc} and MS COCO~\cite{dataset_coco} datasets demonstrate the effectiveness of our methods. The results show that our method boosts the state-of-the-art performance by a large margin.
\end{itemize}

\section{Related Work}
\subsection{Few-shot Learning}
Few-shot learning, which aims to learn general knowledge from data-rich base tasks and quickly adapt to data-poor novel tasks with few labeled examples. The mainstream methods follow the idea of meta-learning, which can be divided into two groups, namely, metric-learning based and optimization based. Metric-learning based approaches, such as MatchNet~\cite{nips16_matchnet} and ProtoNet~\cite{nips17_protonet}, learn a general metric space from base data that adapts well to novel data. Optimization based approaches focus on learning a good initialization point for learning novel data, with representative works including MAML~\cite{icml17_maml} and TAML~\cite{cvpr19_taml}. Besides, some works introduce hallucination techniques to produce synthetic data~\cite{iccv17_halluc}, which alleviates the data-shortage. Recent work in few-shot learning is mostly developed in the context of classification. Compared to few-shot learning, few-shot object detection is more challenging because it consists of two tasks, i.e., classification and localization.

\subsection{Few-shot Object Detection}
Following the idea of few-shot learning, few-shot object detection (FSOD) aims to detect novel categories with abundant base data and scare novel data. Most works follow two types of paradigms, meta-leaning based and fine-tuning based. Meta-learning based works mainly focus on learning class agnostic meta knowledge that can be transferred to novel classes. Meta-YOLO~\cite{iccv19_fsrw} reweights the importance of query features with class-specific support features using channel-wise attention. Attention-RPN~\cite{cvpr20_attentionrpn} is proposed to generate class-relevant proposals and model different relationships between the query and support image. TFA~\cite{icml20_tfa} is the first fine-tuning based method, which reveals the power of freezing the feature extractor and fune-tuning only the last layers. DeFRCN~\cite{iccv21_defrcn} further propose to decouple classification and localization by adjusting back propagation and achieve superior performance. SRR-FSD~\cite{cvpr21_srrfsd} introduces explicit relation reasoning on semantic embeddings, which is robust to the variation of shots for novel objects. Although our work is also fine-tuning based framework that utilizes semantic embeddings. Unlike SRR-FSD, we use semantic embeddings as the unbiased representations for all classes and avoid confusion through our semantic-aware max-margin loss.

\vspace{-1em}
\begin{figure*}[ht]
    \centering
    \includegraphics[width=0.9\textwidth, trim=0 10 0 0, clip]{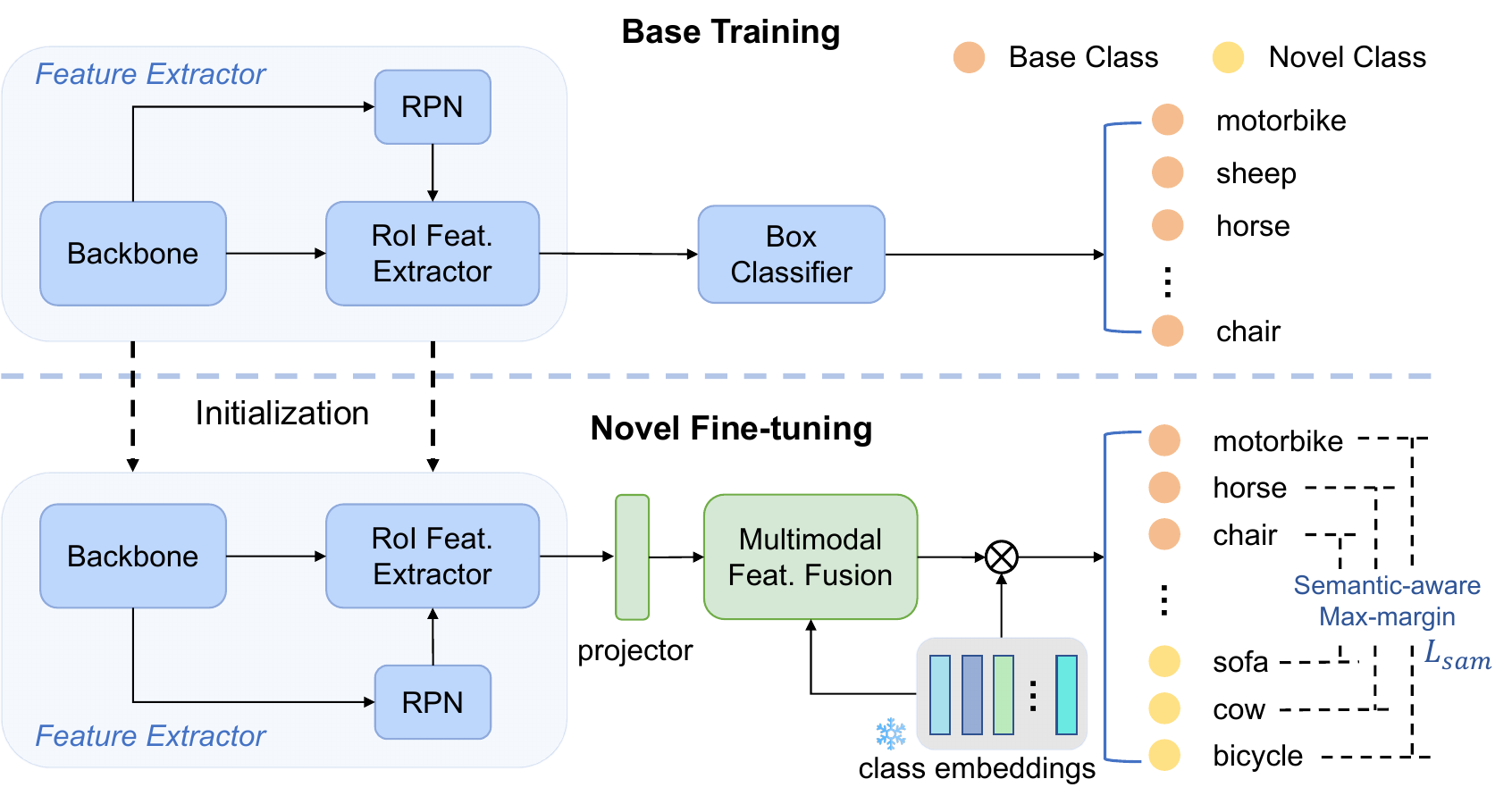}
    \caption{The overview of our method. In the base training stage, we follow previous method to train a linear classifier on the base set. In the novel fine-tuning stage, we initialize the model with base knowledge and replace the linear classifier with our semantic similarity classifier. Additionally, multimodal feature fusion is propose to improve the vision-language communication. Finally the classifier branch is optimized by our semantic-aware max-margin loss.}
    \label{fig:method}

\end{figure*}
\vspace{-0.5em}


\section{Method}
Our work aims to mitigate the low-shot performance degradation and confusion between novel and base classes. The framework of our proposed method is shown in Fig.~\ref{fig:method}. In this section, we first briefly introduce the setting of FSOD problem. Then we present semantic similarity classifier and multimodal feature fusion in Sec.~\ref{subsec:sal} and semantic-aware max-margin in Sec.~\ref{subsec:margin}.

\subsection{FSOD Preliminaries}

In this paper, we follow the few-shot object detection (FSOD) setting as in previous works~\cite{icml20_tfa, iccv21_defrcn}. We divide the training data into a base set $\mathcal{D}_b$ and a novel set $\mathcal{D}_n$, where the base classes $\mathcal{C}_b$ have abundant labeled data while each novel class in $\mathcal{C}_n$ only has a few annotated samples. There is no overlap between base classes and novel classes, namely $ \mathcal{C}_b \cap \mathcal{C}_n = \varnothing $. In the context of transfer-learning, training stages consists of base training on $\mathcal{D}_b$ and novel fine-tuning on $\mathcal{D}_n$. We aim to utilize the generalizable knowledge learned from large base data to quickly adapt to novel classes. The model is expected to detect objects in the test set with classes in $\mathcal{C}_b \cup \mathcal{C}_n$. 

Our approach can be applied to any fine-tuning based few-shot detectors in a plug-and-play fashion and we integrate our approach with the previous state-of-the-art method DeFRCN~\cite{iccv21_defrcn} for verification. DeFRCN~\cite{iccv21_defrcn} is a widely adopted baseline. Different from TFA which freezes most parameters in the second stage to prevent over-fitting, DeFRCN proposes Gradient Decoupled Layer to truncate gradient of RPN and scale gradient of R-CNN in both stages.

\subsection{Semantic Alignment Learning}

\label{subsec:sal}

We aim to utilize the semantic embeddings that provide unbiased representations for all classes to tackle the performance degradation, particularly in extremely low-shot scenarios.

\noindent \textbf{Semantic Similarity Classifier.} Our few-shot detector is built on top of Faster R-CNN~\cite{nips15_fasterrcnn}, a popular two-stage object detector. 
In Faster R-CNN, region proposals are extracted and forwarded to the box classifier and box regressor to generate class labels and accurate box coordinates. Prior fine-tuning based FSOD methods simply expand the classifier with random initialization to generalize to novel categories. However, given only one or two annotated samples of novel objects, the detector can hardly construct an unbiased feature distribution for each novel class, especially when the novel samples are not representative enough. The unbiased feature distributions for novel categories will lead to unsatisfactory detection performance.


To overcome the above obstacle, we propose a semantic similarity classifier and uses the fixed semantic embeddings for recognition instead of the linear classifier. This is based on the observation that the class name embeddings are intrinsically aligned with massive visual information. When training samples are extremely limited, class name embeddings serve as good class centers. 

We first align the region features with the semantic embeddings by a projector, then the cosine similarity between the projected region features and class name embeddings are used to generate the classification scores $\bm{s}$. 
\begin{equation}
\bm{s}=\text{softmax}(\text{D}(\bm{t}, \bm{Pv}))
\label{eq:projection}
\end{equation}
where $\bm{v}$ is the region features, $\bm{P}$ is the projector and $\bm{t}$ is the class name embeddings. $\text{D}$ indicates the distance measurement function.

\noindent \textbf{Multimodal Feature Fusion.} The semantic similarity classifier learns to align the concepts from visual space with the semantic space, but it still treats each class independently and there is no knowledge propagation between modalities except the last layer. This may pose a hindrance to fully exploiting inter-class correlations. Therefore, we further introduce multimodal feature fusion to promote cross-modal communication. The fusion module is based on cross-attention mechanism and conducts aggregation on region features $\bm{v}$ and class name embeddings $\bm{t}$. Mathematically, the process is shown as following.
\begin{equation}
q_v=W^{(q)}\bm{v}, k_t=W^{(k)}\bm{t}, v_t=W^{(v)}\bm{t}
\label{eq:attn_proj}
\end{equation}
\begin{equation}
attention=\text{softmax}(q_vk_t^T/\sqrt{d})
\label{eq:x-attn}
\end{equation}
\begin{equation}
\hat{q_v}=q_v+attention\cdot v_t
\label{eq:residual}
\end{equation}
where {$W^{(q)}, W^{(k)}, W^{(v)}$} are trainable parameters of cross-attention and $d$ is the size of the intermediate channel.

The multimodal fusion module ensures sufficient communication with text features in the early stage of image feature extraction, thus enriching the diversity of region features. Furthermore, it improves the exploitation of inter-class correlation contained in the semantic information. 

\subsection{Semantic-aware Max-margin Loss}
\label{subsec:margin}
The semantic similarity classifier aligns visual feature with semantic embeddings, leading to unbiased feature distributions for novel categories. However, inter-class correlation contained in semantic embeddings may also cause class confusion between similar base and novel classes. To avoid this, we propose semantic-aware max-margin loss which applies an adaptive margin between two classes based on their semantic relation. 

In previous works, the classification branch is optimized by cross entropy loss in an end-to-end way. Each region feature is trained to be close to the class center. Given the $i$-th region feature $v_i$ with label $y_i$, classification loss is computed as follows.
\begin{equation}
L_{cls}=-\frac{1}{n}\sum_{i=1}^{n}\text{log} \ \frac{e^{\text{D}(v_i, t_{y_i})}}{\sum_{j=1}^{n}e^{\text{D}(v_i, t_{y_j})}}
\label{eq:cross_entropy}
\end{equation}
where the $t_{y_i}$ is the class name embedding of $y_i$.

We replace the linear classifier with frozen semantic embeddings. As a result, novel classes can learn from well-trained similar base classes. However, it may also cause confusion if the semantic relation between the two categories is very close. Therefore, we add an adaptive margin onto the cross entropy loss and push confusable classes away from each other. Mathematically the semantic-aware max-margin loss are calculated as follows.
\begin{equation}
L_{sam}=-\frac{1}{n}\sum_{i=1}^{n}\text{log} \ p_i
\label{eq:max-margin1}
\end{equation}
where $p_i$ indicates the classification score,
\begin{equation}
p_i=\frac{e^{\text{D}(v_i, t_{y_i})}}{e^{\text{D}(v_i, t_{y_i})} + \sum_{j\neq i}^{n}e^{\text{D}(v_i, t_{y_j}) + m_{ij}}}
\label{eq:max-margin2}
\end{equation}
in which $m_{ij}$ indicates the margin applied between class $i$ and class $j$,
\begin{equation}
m_{ij}=
\begin{cases}
\text{cosine}(t_i, t_j)& \text{cosine}(t_i, t_j)-\gamma>0 \\
0& \text{cosine}(t_i, t_j)-\gamma\leq0
\end{cases}
\label{eq:max-margin3}
\end{equation}
where $\gamma$ is the threshold for semantic similarity. For each class, we choose only the top $k$ most similar classes to apply the margins to avoid unnecessary noise.
\begin{table*}[htbp]
    \centering
    \small
\caption{FSOD results (\%) on Pascal VOC. We report comparison over nAP50. $\dag$ indicates results reproduced by us. The {\bf best} and \ul{second-best} are highlighted.}

\begin{tabular}{l|ccccc|ccccc|ccccc}
\toprule
\multirow{2}{*}{Method / Shots}  & \multicolumn{5}{c|}{Novel Split 1} & \multicolumn{5}{c|}{Novel Split 2} & \multicolumn{5}{c}{Novel Split 3}  \\
                   & 1            & 2            & 3            & 5            & 10           & 1            & 2            & 3            & 5            & 10           & 1            & 2            & 3            & 5            & 10       \\ 
\midrule
FSRW~\cite{iccv19_fsrw}               & 14.8         & 15.5         & 26.7         & 33.9         & 47.2         & 15.7         & 15.3         & 22.7         & 30.1         & 40.5         & 21.3         & 25.6         & 28.4         & 42.8         & 45.9     \\
Meta R-CNN~\cite{iccv19_metarcnn}         & 19.9         & 25.5         & 35.0         & 45.7         & 51.5         & 10.4         & 19.4         & 29.6         & 34.8         & 45.4         & 14.3         & 18.2         & 27.5         & 41.2         & 48.1     \\
TFA w/ cos~\cite{icml20_tfa}       & 39.8         & 36.1         & 44.7         & 55.7         & 56.0         & 23.5         & 26.9         & 34.1         & 35.1         & 39.1         & 30.8         & 34.8         & 42.8         & 49.5         & 49.8      \\
CME~\cite{cvpr21_cme}                & 41.5         & 47.5         & 50.4         & 58.2         & 60.9         & 27.2         & 30.2         & 41.4         & 42.5         & 46.8         & 34.3         & 39.6         & 45.1         & 48.3         & 51.5      \\
SRR-FSD~\cite{cvpr21_srrfsd}        & 47.8         & 50.5         & 51.3         & 55.2         & 56.8         & 32.5         & 35.3         & 39.1         & 40.8         & 43.8         & 40.1         & 41.5         & 44.3         & 46.9         & 46.4      \\
FSCE~\cite{cvpr21_fsce}                & 44.2         & 43.8         & 51.4         & 61.9         & 63.4         & 27.3         & 29.5         & 43.5         & 44.2         & 50.2         & 37.2         & 41.9         & 47.5         & 54.6         & 58.5      \\
QA-FewDet~\cite{iccv21_qafewshot}          & 42.4         & 51.9         & 55.7         & 62.6         & 63.4         & 25.9         & 37.8         & 46.6         & 48.9         & 51.1         & 35.2         & 42.9         & 47.8         & 54.8         & 53.5      \\
FADI~\cite{nips21_fadi}                 & 50.3         & 54.8         & 54.2         & 59.3         & 63.2         & 30.6         & 35.0         & 40.3         & 42.8         & 48.0         & 45.7         & 49.7         & 49.1         & 55.0         & 59.6      \\
Meta FR-CNN\cite{aaai22_metafaster}        & 43.0 & 54.5 & 60.6 & 66.1 & 65.4 & 27.7 & 35.5 & 46.1 & 47.8 & 51.4 & 40.6 & 46.4 & 53.4 & 59.9 & 58.6     \\
VFA~\cite{aaai23_vfa}               & \ul{57.7} & \ul{64.6} & 64.7 & 67.2 & 67.4 & \bf{41.4} & \ul{46.2} & 51.1 & 51.8 & 51.6 & \ul{48.9} & 54.8 & 56.6 & 59.0 & 58.9   \\
\midrule
DeFRCN$\dag$~\cite{iccv21_defrcn}   & 55.2	& 64.3	& \ul{65.7}	& \ul{68.4}	& \ul{69.0}	& 33.9	& 45.8	& \ul{51.5}	& \ul{54.1}	& \bf{53.2}	& 48.8	& \ul{54.0}	& \bf{59.4}	& \bf{62.9}	& \bf{63.5}  \\
\bf{Ours}   & \bf{58.7}	& \bf{67.2}	& \bf{68.5}	& \bf{69.8}	& \bf{69.8}	& \ul{35.9}	& \bf{46.8}	& \bf{53.8}	& \bf{55.9}	& \ul{52.9}	& \bf{52.3}	& \bf{55.8}	& \ul{59.1}	& \ul{62.4}	& \ul{62.6} \\
\bottomrule
\end{tabular}

\label{Table:result-voc-sota}
\end{table*}

\begin{table}[htb]
    \centering
    \small
\caption{FSOD results (\%) on COCO. We report comparison over COCO-style nAP and nAP75. $\dag$ indicates results reproduced by us. The {\bf best} and \ul{second-best} are highlighted.}

\begin{tabular}{l|cc|cc}
\toprule
\multirow{2}{*}{Method / Shots} & \multicolumn{2}{c|}{10} & \multicolumn{2}{c}{30}                                                                 \\
                         & \multicolumn{1}{c}{AP} & \multicolumn{1}{c|}{AP75} & \multicolumn{1}{c}{AP} & \multicolumn{1}{c}{AP75} \\ \midrule

Meta R-CNN~\cite{iccv19_metarcnn}   & 8.7 & 6.6 & 12.4 & 10.8 \\
TFA w/ cos~\cite{icml20_tfa}                          & 10.0 & 9.3 & 13.7 & 13.4 \\
SRR-FSD~\cite{cvpr21_srrfsd}                             & 11.3 & 9.8 & 14.7 & 13.5 \\
FSCE~\cite{cvpr21_fsce}                                & 11.1 & 9.8 & 15.3 & 14.2 \\
CME~\cite{cvpr21_cme}                                 & 15.1 & 16.4 & 16.9 & 17.8 \\
QA-FewDet~\cite{iccv21_qafewshot}                           & 11.6 & 9.8 & 16.5 & 15.5 \\
FADI~\cite{nips21_fadi}                                & 12.2 & 11.9 & 16.1 & 15.8 \\                         
Meta FR-CNN~\cite{aaai22_metafaster}                         & 12.7 & 10.8 & 16.6 & 15.8 \\
VFA~\cite{aaai23_vfa}                                 & 16.2 & -    & 18.9 & -    \\  \midrule
DeFRCN$\dag$~\cite{iccv21_defrcn}                       & \ul{19.0} & \bf{18.6} & \ul{22.2} & \bf{22.2} \\
Ours                                & \bf{20.0} & \ul{18.5} & \bf{22.6} & \ul{21.9} \\

\bottomrule
\end{tabular}

\label{Table:result-coco-sota}
\end{table}


\section{Experiments}

\subsection{Experimental Setting}

\noindent \textbf{Existing benchmarks.} 
Following previous works~\cite{iccv19_fsrw,icml20_tfa, iccv21_defrcn}, we use Pascal VOC~\cite{dataset_voc} and MS COCO~\cite{dataset_coco} to evaluate our methods. For Pascal VOC, we split the overall 20 classes into 15 base classes and 5 novel classes. We consider three different partitions and refer them as Novel Split 1, 2, 3 respectively. Each base class contains abundant data while every novel class has only $K$ annotated instances with $K = 1,2,3,5,10$. As for MS COCO, we set the 60 classes disjoint with VOC as base classes and the other 20 classes as novel classes. We use $K = 1,2,3,5,10,30$ shots settings for COCO.

\noindent \textbf{Evaluation setting.}
For Pascal VOC, we report the AP50 of novel classes (nAP50) on VOC07 test set. For COCO, we report COCO-style mAP of novel classes (nAP) on COCO 2014 validation set.

\noindent \textbf{Implementation details.} 
We implement our approach with mmdetection~\cite{mmdetection}. Faster-RCNN is our basic detection framework and ResNet-101~\cite{cvpr16_resnet} pre-trained on ImageNet~\cite{ijcv15_imagenet} is used as the backbone. We train our detector in a multi-task fashion, where the localization branch and semantic alignment learning branch are optimized simultaneously. We generate our text embeddings using CLIP~\cite{icml21_clip} as the text encoder and class names as the prompt. Each word embedding is L2-normalized along the feature dimension. The similarity threshold in max-margin loss is set as $0.5$ empirically. We keep other hyper-parameters the same as DeFRCN~\cite{iccv21_defrcn}.

\subsection{Comparison Results}

\noindent \textbf{PASCAL VOC.} 
Our evaluation results of Pascal VOC on three different splits are shown in Table~\ref{Table:result-voc-sota}. Our method outperforms existing approaches on most settings~(10/15) and achieves second-best~(5/15) for the rest. In Novel set 1, our approach exceeds the previous SOTA by 0.8$\sim$2.8 mAP. Even though DeFRCN~\cite{iccv21_defrcn} is a pretty strong baseline, our method are able to further boost the performance by a large margin, especially for extremely low-shot settings ($shots\leq3$). Our method achieves 6.3\%, 5.9\% and 7.2\% improvement on 1-shot setting for three splits respectively. In addition, the 2-shot results for all three settings and 3-shot results for split 1 and 2 show similar improvement. It is reasonable because semantic concepts play a more important role when visual information become scarce. However, in Novel set 3, our method is slightly behind DeFRCN as the shot number grows. We conjecture that semantic information could confuse the model when visual samples are sufficient.

\noindent \textbf{COCO.} 
Table~\ref{Table:result-coco-sota} shows all evaluation results on COCO with COCO-style averaged AP~($mAP$). Our approach achieves the best nAP under 10-shot and 30-shot settings, surpassing the previous SOTA by 5.3\% and 1.8\% respectively. Compared to DeFRCN, our method shows stronger robustness and generalization ability. It also shows that our approach continues to hold up on challenging datasets.

\begin{figure}[t]
    \centering
    \includegraphics[width=\linewidth]{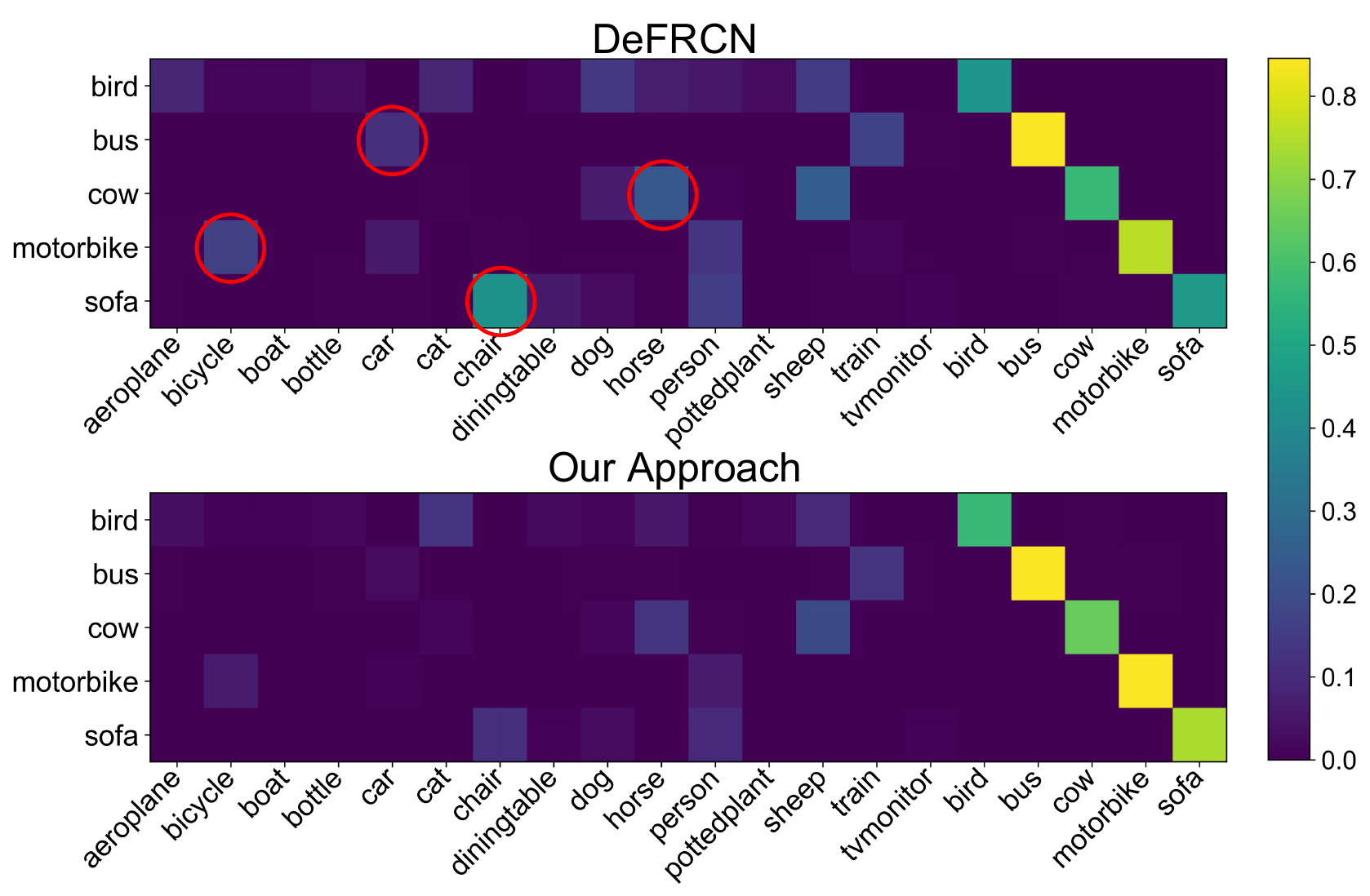}
    \caption{The confusion matrix on Pascal VOC Split1. Each element in column~$m$ and row~$n$ indicates the percentage of samples in class $m$ that are recognized as class $n$. If $m$ and $n$ stand for different classes, a high score would indicate severe confusion. Our approach alleviates the confusion between novel classes and similar base classes by a large margin.}
    \label{fig:confusion}
\end{figure}

\subsection{Ablation Studies}

In this section, we perform a detailed ablation study of each component of our method. We first investigate the effectiveness of each module in our model by gradually applying our proposed components to DeFRCN~\cite{iccv21_defrcn}. Then we explore the effect of using different intermediate channels for MFF and varying the number of most similar classes in SAM loss. All experiments are conducted on VOC Novel Set 1. 

\begin{table}[t]
    \centering
    \small
    \caption{Effectiveness of different modules. Ablative performance~(nAP50) on the VOC novel set 1 is reported. $\dag$ indicates results reproduced by us. The {\bf best} and \ul{second-best} are highlighted. {\bf SSC}: Semantic Similarity Classifier. {\bf MFF}: Multimodal Feature Fusion. {\bf SAM}: Semantic-Aware Max-margin.}

    \begin{tabular}{lccc|ccc}
    \toprule
    \multirow{2}{*}{Method}     & \multirow{2}{*}{SSC}       & \multirow{2}{*}{MFF}       & \multirow{2}{*}{SAM}            & \multicolumn{3}{c}{Shots}  \\
                                &                           &                            &                           & 1        & 2        & 3                       \\ \midrule 
    \multicolumn{1}{l}{DeFRCN}  &                           &                            &                           & 55.2        & 64.3        & 65.7            \\ \midrule
    \multirow{3}{*}{Ours}       & \checkmark                &                            &                           & 55.8        & 66.5        & 67.5            \\
                                & \checkmark                & \checkmark                 &                           & 58.2        & 66.6        & 67.7            \\
                                & \checkmark                &                            & \checkmark                & 55.9        & \bf{67.3}        & 68.2            \\          
                                & \checkmark                & \checkmark                 & \checkmark                & \bf{58.7}   & 67.2   & \bf{68.5}           \\ \bottomrule
    \end{tabular}
    \label{Table:ablations}

\end{table}

\begin{figure}[t]
    \centering
    \includegraphics[width=\linewidth, trim=0 15 0 0, clip]{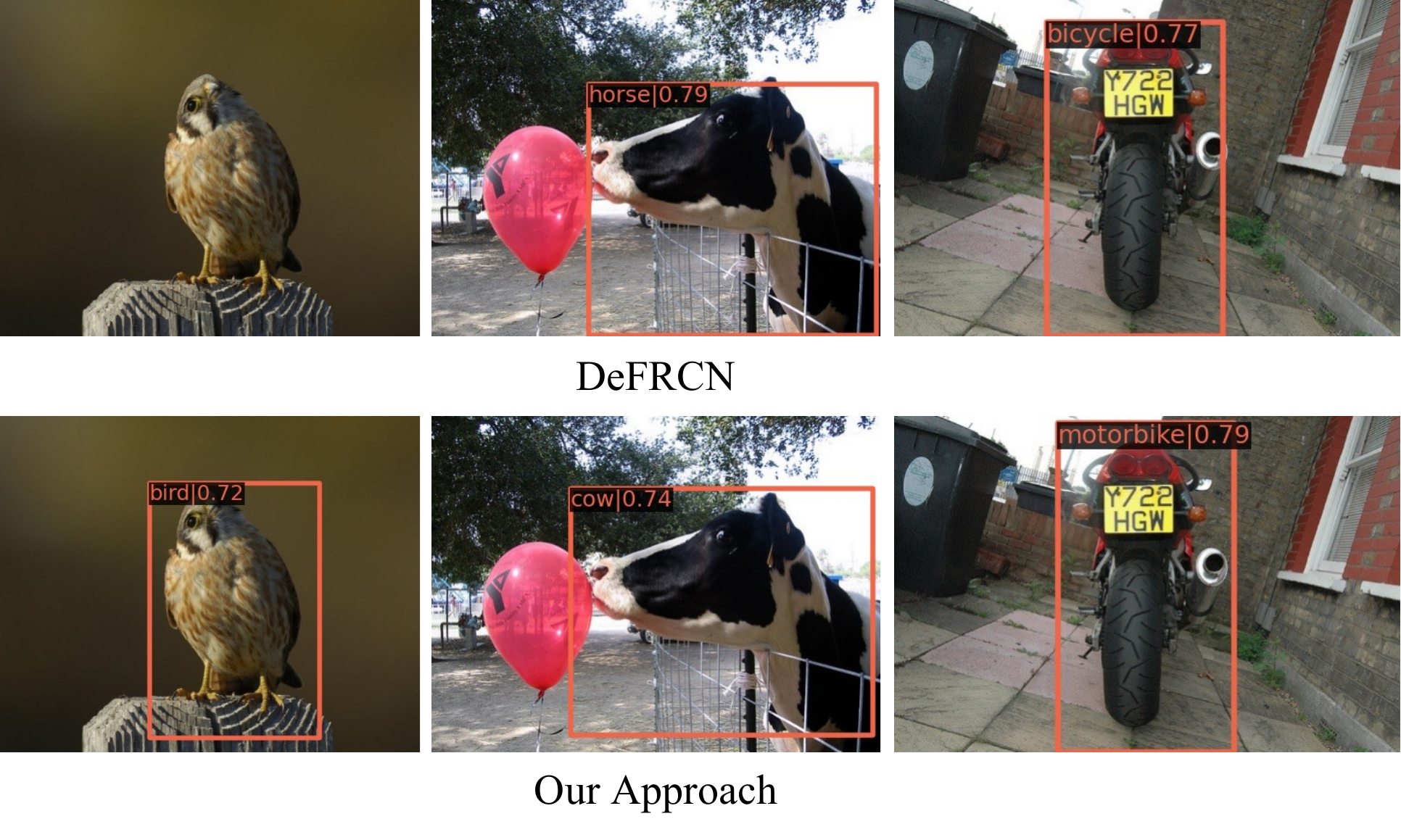}
    
    \caption{Visualization of detection results on the VOC Split1. Our method detects objects that DeFRCN misses~(left) and confuses~(middle and right).}
    
    \label{fig:comparison_visualization}
\end{figure}

\begin{table}[t]
    \centering
    \small
    \caption{Ablation on the intermediate channel dimension of multimodal feature fusion.}

    \begin{tabular}{c|ccccc}
    \toprule   
    \multirow{2}{*}{Dimension}           &   \multicolumn{5}{c}{Shots in Novel Set 1}                              \\                         
                                        & 1        & 2        & 3       & 5     & 10                \\ \midrule 
    512                         & 58.0	& 67.3	& 67.7	& 68.6	& 69.4            \\                      
    256                         & 58.4	& 67.2	& 67.7	& 68.5	& 69.7            \\ 
    128                         & 58.3	& 67.2	& 67.9	& 68.4	& 69.6            \\
    64                          & 57.8	& 67.3	& 67.7	& 69.3	& 69.1             \\
    \rowcolor[gray]{.8} 32      & \bf{58.7}	& 67.2	& \bf{68.5}	& \bf{69.8}	& \bf{69.8}            \\          
    16                          & 57.9	& \bf{67.7}	& 68.1	& 68.6	& 68.8           \\ \bottomrule
    \end{tabular}
    \label{Table:inter_channels}

\end{table}

\begin{table}[htb]
    \centering
    \small
    \caption{Ablation on the number of similar classes $k$.}

    \begin{tabular}{c|ccccc}
    \toprule   
    \multirow{2}{*}{$K$}           &   \multicolumn{5}{c}{Shots}                              \\                         
                                        & 1        & 2        & 3       & 5     & 10                \\ \midrule 
    1                      & 58.5	& 66.8	& 67.9	& 68.6	& 69.2            \\ 
    2                      & 57.9	& 67.1	& 67.8	& 68.7	& 69.7            \\
    \rowcolor[gray]{.8} 3                     & \bf{58.7}	& \bf{67.2}	& \bf{68.5}	& \bf{69.8}	& \bf{69.8}            \\
    4 & 57.7	& 66.6	& 67.5	& 68.9	& 69.3            \\          
    all                     & 58.2	& \bf{67.2}	& 68.0	& 68.5	& 69.0           \\ \bottomrule
    \end{tabular}
    \label{Table:ablation_ksim}
\vspace{-1.0em}
\end{table}

\noindent \textbf{Effectiveness of different modules.}
We conduct relative ablations to analyze how much each module contributes to our method in Table~\ref{Table:ablations}. As the baseline already achieve satisfying performance at 5-shot and 10-shot, the results of low-shot scenarios (i.e., 1-shot, 2-shot and 3-shot) are used for comparison. Although DeFRCN is competitive enough, we demonstrate significant improvement in its performance. Replacing the linear classifier with a cosine similarity based classifier and incorporating MFF have boosted the performance by a large margin on all shot settings. It is reasonable, as the class name embeddings are assigned as unbiased representations for novel classes when the shot is extremely low. In Table~\ref{Table:ablations}, we find that SSC has a relatively minor improvement in the 1-shot scenario.  
We conjecture that over-fitting in the projector is the reason for the discrepancy. With the help of MFF, the only one region feature for each class will interact with semantic embeddings to become more diversified. Besides, the SAM loss also shows robust improvement in all shot scenarios.

\noindent \textbf{Intermediate channel dimension of multimodal feature fusion. } 
Each region feature is projected to the semantic space and interacted with class name embeddings in the MFF module. We investigate the effect of intermediate channel dimension in the cross-attention. It can be drawn that as the channel dimension grows, the module is prone to overfit due to more parameters. As the results shown in Table.~\ref{Table:inter_channels}, 32 is the most suitable intermediate channel dimension and is sufficient to support communication between two modals.

\noindent \textbf{$\bm{K}$ in Semantic-aware Max-margin loss. } 
In the max-margin loss, we add an adaptive margin $m_{ij}$ to the Cross Entropy loss based on semantic similarity between class $i$ and class $j$. For each region proposal predicted as class~$i$, we choose the $k$ most similar classes and apply their corresponding margin. We conduct ablation on the effect of $k$ and the results are shown in Table.~\ref{Table:ablation_ksim}. Either too high or too low is unsuitable for the choice of $k$. It is reasonable that a small $k$ might not provide sufficient semantic relation and a large k may bring in too much noise and affect the training process. Although there is a difference on the number of classes between VOC and COCO, we experimentally find that 3 is also the most suitable $k$ for COCO.

\noindent \textbf{Visualization.} 
To show the effectiveness of our method, we visualize the confusion matrix on VOC Split~1 in Fig.~\ref{fig:confusion}. Compared with DeFRCN, our approach relieves the confusion between novel categories and their similar base categories by a large margin. As shown in Fig.~\ref{fig:comparison_visualization}, our method achieves better localization and classification performance.

\vspace{-0.5em}
\section{Conclusion}
\vspace{-0.5em}

In this paper, we propose a FSOD framework that effectively exploits the semantic information reflected in the class name. The semantic similarity classifier learns to align region features with class name embeddings and the multimodal feature fusion promote information exchange between two modalities. The semantic-aware max-margin loss alleviates confusion by push similar novel and base classes away from each other. Experiments results demonstrate that our approach boost the performance of the previous best FSOD method, especially in low-shot scenarios.



\bibliographystyle{IEEEbib}
\bibliography{refs}

\end{document}